\documentclass[lettersize,journal]{IEEEtran}
\usepackage{amsmath,amsfonts}
\usepackage{algorithmic}
\usepackage{algorithm}
\usepackage{array}
\usepackage[caption=false,font=normalsize,labelfont=sf,textfont=sf]{subfig}
\usepackage{textcomp}
\usepackage{stfloats}
\usepackage{url}
\usepackage{verbatim}
\usepackage{graphicx}
\usepackage{cite}

\usepackage{latexsym}
\usepackage{amssymb}
\usepackage{amsthm}
\usepackage{booktabs}
\usepackage{enumitem}
\usepackage{graphicx}
\usepackage{color}
\usepackage{multirow}
\usepackage{textcomp}
\usepackage{enumitem} 
\usepackage{afterpage}  

\usepackage{subfig}
\usepackage{float} 
\usepackage{caption} 
\usepackage{xcolor}

\begin{document}

\title{Semantic Frame Aggregation-based Transformer for Live
Video Comment Generation}

\author{Anam Fatima, 
       Yi Yu*,~\IEEEmembership{Senior Member,~IEEE, }%
       Janak Kapuriya,
       Julien Lalanne
       and Jainendra Shukla 
\thanks{*Corresponding author.}
\thanks{A. Fatima, J. Kapuriya and J. Shukla are with IIIT-Delhi, India.\\
Y. Yu is with Graduate School of Advanced Science and Engineering at Hiroshima University. J. Lalanne is with Grenoble INP – Ensimag, France. (e-mail: anamf@iiitd.ac.in, yiyu@hiroshima-u.ac.jp, julien.lalanne.travail@gmail.com, kapuriya22032@iiitd.ac.in, jainendra@iiitd.ac.in)}} 

\maketitle
\begingroup
\renewcommand\thefootnote{}
\footnotetext{\textbf{Preprint Notice:} This is the author’s preprint version of the article published in \emph{IEEE Transactions on Multimedia (TMM)}, 2025. Final version: \url{https://doi.org/10.1109/TMM.2025.3604921}.}
\endgroup
\begin{abstract}
Live commenting on video streams has surged in popularity on platforms like Twitch, enhancing viewer engagement through dynamic interactions. However, automatically generating contextually appropriate comments remains a challenging and exciting task. Video streams can contain a vast amount of data and extraneous content. Existing approaches tend to overlook an important aspect of prioritizing video frames that are most relevant to ongoing viewer interactions. This prioritization is crucial for producing contextually appropriate comments. To address this gap, we introduce a novel Semantic Frame Aggregation-based Transformer (SFAT) model for live video comment generation. This method not only leverages CLIP’s visual-text multimodal knowledge to generate comments but also assigns weights to video frames based on their semantic relevance to ongoing viewer conversation. It employs an efficient weighted sum of frames technique to emphasize informative frames while focusing less on irrelevant ones. Finally, our comment decoder with cross-attention mechanism to attend to each modality ensures that the generated comment reflects contextual cues from both chats and video. Furthermore, to address the limitations of existing datasets, which predominantly focus on Chinese-language content with limited video categories,, we have constructed a large-scale, diverse, multimodal English video comments dataset. Extracted from Twitch, this dataset covers $11$ video categories, totaling $438$ hours and $3.2$ million comments. We demonstrate the effectiveness of our SFAT model by comparing it to existing methods for generating comments from live video and ongoing dialogue contexts.
\end{abstract}


\section{Introduction}
\IEEEPARstart{L}{ive} commenting on videos has become a popular feature in live streaming platforms such as Twitch, YouTube, Bilibili, Facebook and Instagram. Also known as ``bullet screen" or ``danmaku", it offers a dynamic and interactive experience, promoting engagement and conversations among viewers \cite{livebot, knowledge, zhang2023vcmaster}. In contrast to traditional video comments, which neither reference specific moments in the video nor interact with one another, danmaku comments enable rich multimodal information interactions \cite{videoic}. It fosters a dynamic multimodal group-chatting conversational experience involving more than two speakers. 

Developing multimodal conversational agents capable of engaging in relevant and coherent interactions has been a long-pursued objective in Artificial Intelligence (AI) \cite{pasunuru-bansal-2018-game}. Inspired by it, automatic live commenting is being explored as a challenging testbed for AI agents requiring simultaneous understanding of multimodal contexts from live streams and ongoing viewer conversations. It poses greater challenges compared to other multimodal interaction tasks such as video captioning \cite{Zhou_2018_CVPR, Lin_2022_CVPR, Gu_2023_CVPR} or visual question-answering \cite{lei2018tvqa, le2020hierarchical, Yang_2020_WACV}. Video comments involve diverse user conversations, subjective opinions, reactions, and discussions that can sometimes diverge from the live-streaming video. In contrast, video captioning focuses on objective description of the visual content, while video question-answering involves retrieving factual information from the video itself. 

\begin{figure}[!t]
  \centering
  \includegraphics[width=0.50\textwidth]{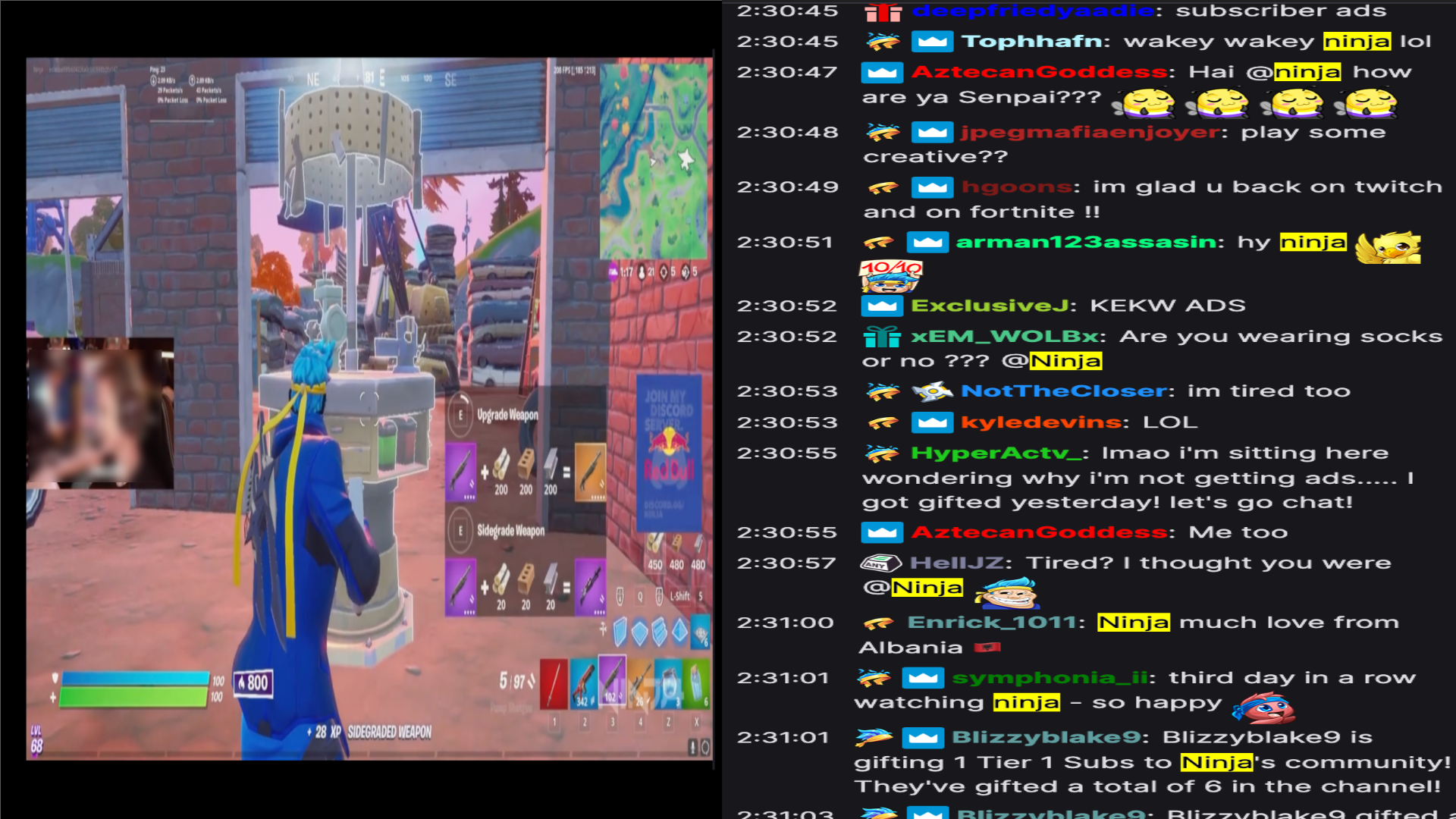}
  \caption{
  An example of a live video with viewer chats from Twitch, illustrating why essential to align visual elements with chat references for contextual relevance.
  }
  \label{fig:1}
\end{figure}

We see that automatically generating video comments is an exciting yet challenging domain, as evidenced by ongoing research efforts \cite{livebot, knowledge, zhang2023vcmaster, dca, luo2024engaging}. However, existing methods \cite{livebot, knowledge, videoic, plvcg, coldstart, zhang2023vcmaster} overlook an important aspect: the prioritization of video frames that are pivotal to ongoing conversations. Assigning relevance to frames based on contextual comments ensures that the generated content reflects the true interests of viewers, aligning with their engagement. For instance, in Figure \ref{fig:1}, a Minecraft gameplay scene is shown alongside a highly active chat, discussing gameplay observations with multiple references to ``Ninja” (highlighted in chats). This highlights the complex multimodal dependencies in such contexts, where comments are often reactions to the video or discussions among viewers. To enhance viewer engagement, it is essential for the model to prioritize and align the visual elements with these chat references (e.g., the ``Ninja" avatar and its actions), ensuring the generated comments are contextually relevant to what viewers are discussing.

Current approaches fail to address this by treating all video frames equally, without adequately prioritizing those most relevant to ongoing viewer interactions. This lack of prioritization often leads to generic and less engaging comments that do not capture the essence of the conversation. By prioritizing frames aligned with the conversational context, the model can generate targeted, accurate, and context-aware comments (such as ``what is Ninja doing next?") that resonate with viewer interests. To bridge this gap, our work introduces a novel technique within the live video comment generation domain to assign weights to video frames based on the ongoing dialogue context. We propose the \textbf{Semantic Frame Aggregation-based Transformer (SFAT)}\footnote{Our code and processed dataset will be publicly available on GitHub. Raw data can be provided upon request.} model to facilitate the extraction and integration of live video and context comments modalities to generate contextually coherent comment. Unlike prior works \cite{livebot, coldstart, plvcg, videoic, zhang2023vcmaster} that employ standard multimodal encoder under-utilizing interdependence of modalities, our approach introduces an intricate multimodal encoder employing sophisticated weighting mechanism to prioritize frames based on chat-context. It leverages CLIP’s visual-text multimodal knowledge to enhance video content representation and combines information from a series of video frames based on semantic similarity to the context comments. Thus, it efficiently utilizes both context comments and video frames to generate relevant and coherent comments.  To demonstrate the efficacy of our proposed SFAT approach, we perform a comparative study with the Triple Transformer Encoder (TTE) model \cite{livebot}, its variants \cite{resnet, clip}, VideoIC \cite{videoic}, KLVCG \cite{knowledge} and state-of-the-art multimodal LLM Video-ChatGPT \cite{maaz2023video}.

Most of the existing multimodal video-chat datasets \cite{livebot, coldstart, zhang2023vcmaster} predominantly focus on the \textbf{Chinese} language content often sourced from platforms like Bilibili or the MovieLC dataset \cite{knowledge} from a Chinese movies platform, covering limited categories and lacking diversity. This underscores a clear need for comprehensive multimodal video comment datasets in other global languages catering to a broader audience, such as \textbf{English}, to support the advancement of live commenting technologies. To address this, we create a large-scale, diverse, multimodal English live-video comment dataset with popular categories from Twitch to help researchers and practitioners with advanced research in English content. Twitch is one such platform supporting content in multiple languages, including English. It has evolved beyond gaming content to include diverse categories like music, art, talk shows, and more, attracting a variety of content creators and viewers. Our \textbf{VideoChat} dataset, extracted from Twitch, comprises $11$ categories and $575$ streamers totaling $438$ hours of video and $3.2$ million comments. The task becomes more challenging for English due to diverse global accents and dialects. English chats on live-streaming platforms exhibit significant linguistic diversity, including informal language, unstructured and spontaneous reactions, slang, gaming lingos, abbreviations, and regional variations. Training models on such a challenging dataset for English has been largely unexplored in previous studies, establishing it as a more demanding and impactful benchmark.

Beyond video-comment generation, in the future, such English multimodal video-chat datasets can be extended for other related contextual understanding tasks like video-grounded conversational AI, dialogue-driven video retrieval, and keyframe extraction. They can be instrumental in developing video-grounded applications, such as personal assistants, human-computer interaction systems, and intelligent tutoring systems \cite{pasunuru-bansal-2018-game} for English-speaking users. Unlike traditional NLP datasets with structured text and correct grammar, live-chat datasets feature informal language, brief reactive messages, and grammatical errors \cite{10.5555/3505464.3505501}. Such a dataset can be extended for advancing NLP research in informal language understanding and generation as well as multimodal temporal sentiment analysis and emotion detection based on viewer's reactions.

To this end, main contributions of our work are as follows:
\begin{enumerate} [label=\roman*)]
  
  \item Facilitating the prioritization of keyframes by proposing a novel \textbf{Semantic Frame Aggregation-based Transformer (SFAT)} model integrating video-frames and context comments to generate contextually coherent comment. It employs two primary techniques: extraction of video embeddings through semantic frame aggregation and an effective multimodal encoder. 
  
  \item Intricately designing our \textbf{multimodal encoder} to extract and combine visual and contextual dialogue modalities effectively based on the weighted sum mechanism. It leverages the CLIP’s \cite{clip} multimodal knowledge to assign weights to video frames based on their semantic similarity to ongoing dialogue.

  \item Integrating the \textbf{comment decoder} with modality-specific cross-attention mechanism that not only aligns the masked target comment with contextual comment embeddings from the text encoder but also incorporates visual information from the aggregated video frame embeddings. This ensures that the generated comment reflects contextual cues from both chats and video.
  
  \item Constructing \textbf{VideoChat}, a large-scale, diverse, multimodal video content dataset extracted from Twitch in \textbf{English} language, thus catering to a wider global audience and advancing research in English content.
  
  \item Through extensive experiments on this dataset, we demonstrate the efficacy of the SFAT model in leveraging visual-textual contexts. It lays the groundwork for future research with potential applications in related research tasks, extending beyond video-comment generation.
  
\end{enumerate}

\section{Related Work}
Live comment generation shares some similarities with other video analysis tasks such as video captioning \cite{Zhou_2018_CVPR, Lin_2022_CVPR, Gu_2023_CVPR}, visual question-answering \cite{lei2018tvqa, le2020hierarchical, Yang_2020_WACV} and video summarization \cite{huang2021gpt2mvs, narasimhan2021clip, zhao2022hierarchical}, involving video content understanding to generate textual information. However, automatic live video comment generation has unique challenges in handling multi-user interactions concurrently with ongoing dynamic video streams. The user interactions can span over diverse topics and styles, each characterized by unique terminology, including slang, emojis, and abbreviations. In contrast, the scope of analysis for other multimodal tasks \cite{Lin_2022_CVPR, Yang_2020_WACV, huang2021gpt2mvs} is usually confined to a question or providing objective information or descriptions strictly based on the video content. Furthermore, the language in questions, summaries, and captions is typically more structured and restricted compared to the broad spectrum of expressions found in live comments.

\subsection{Existing Methods and Research Gap}
The challenging task has been introduced by LiveBot \cite{livebot}, where the authors proposed a ``bullet screen" benchmark dataset extracted from Bilibili. It is the pioneering work providing a benchmark for generating live-video comments based on multimodal inputs. It proposed a Long Short-Term Memory (LSTM) \cite{lstm} based model for comparison and a unified multimodal Transformer \cite{attention-is-all-you-need} for the main architecture. It also introduced retrieval-based metrics appropriate for this task, asking the model to sort candidate comments set based on the log-likelihood score. Subsequent works have addressed various challenges and introduced improvements to this work. In the paper Response to LiveBot \cite{response_livebot}, the authors highlighted some shortcomings in the earlier work, leading to further investigations into optimizing the task's results. 

Some other works, such as VideoIC \cite{videoic} and \cite{wu2021knowing}, worked on leveraging the temporal relation between interactions for comment generation. VideoIC employed a multimodal multitask learning-based approach for comment generation. The multimodal encoding included a video encoder for generating both global and local representation for video frames and a text encoder for encoding surrounding comments. For multitask learning, in addition to the comment generation task, they introduced a temporal relation prediction task to predict the relation between the current time-stamp and the target time-stamp, whether in the left context or the right context of the target. However, this method requires access to both past and future contexts for generating comments and, hence, is not suitable for live-commenting tasks having access to past chat-context only.

Notably, KLVCG \cite{knowledge} used a knowledge-enhanced approach to generate contextually rich comments for a newly constructed long video dataset, MovieLC \cite{knowledge}. However, this method making use of external knowledge \cite{knowledge_graph} is well-suited to movie datasets where factual information about characters, plots, and settings is crucial for generating contextually rich comments and may be less effective for diverse datasets like ours, lacking fixed narrative structures or involving spontaneous interactions, such as those found in gaming content.

VCMaster \cite{zhang2023vcmaster} aimed at generating diverse comments without repetitive words by enlarging the distance between generated and contextual comments using Sentence-Level Contrastive Loss. It used diversity-based metrics, such as distinct and contrastive loss-based scores, to measure the variety in generated comments instead of retrieval-based metrics, as used in LiveBot. While VCMaster focuses on comment diversity relative to available context comments, our work targets to generate comments aligned with the ongoing viewer interactions by prioritizing informative frames.  CMVCG \cite{9533460} used a non-autoregressive approach with a keywords position predicting module to generate comments. It relies on user-provided keywords prompt to guide comment generation, while our approach aims to generate coherent comments without user guidance.

Engaging Live Video Comments Generation \cite{luo2024engaging} aimed to generate comments that attract audience interaction by maximizing “like” counts. It employs contrastive learning loss (Semantic Gap Contrastive Loss) to push the generated comments closer to highly-liked examples and farther from lower-liked comments. 
It uses a dataset explicitly annotated with ``like" counts. It may not be applicable to our dataset extracted from the Twitch platform, where ``like" counts are not available for live-streaming videos. Another work, \cite{coldstart}, focused on comment generation strategy for less-commented videos, training a multimodal transformer for different comment densities. PLVCG \cite{plvcg} incorporated comment posted-time and video-label information for improved comments generation.

These works aimed at introducing new methodologies to capture temporal relations \cite{videoic}, improve sentence level diversity \cite{zhang2023vcmaster}, enhance knowledge \cite{knowledge} or use user-prompts \cite{9533460} for comment generation task function. Some of them focused on improving datasets by adding information such as ``like" counts \cite{luo2024engaging} for user-guided generation or incorporating posted-time \cite{plvcg}. As discussed above, none of the existing works address the prioritization of informative video-frames relative to chat-context to generate contextually relevant comment. The existing works particularly exploit standard multimodal encoders for video and comments representation, under-utilizing interdependence of modalities and alignment of relevant visual features with dialogue context. We intricately designed our multimodal encoder leveraging pretrained CLIP \cite{clip} to obtain semantically-aggregated frame embeddings relative to chat-context for contextually relevant comment generation and enhanced user-engangement.
Furthermore, our comment decoder with a modality-specific cross-attention mechanism, not only aligns the masked target comment with the context comment embeddings from the text encoder—allowing it to incorporate textual cues from the chat context—but also integrates information from the aggregated video frame embeddings, ensuring the generated comment accurately reflects the visual context.

\subsection{Limitations of Existing Datasets}


Most datasets for live video comment generation are limited in scope and diversity. Prominent datasets, such as those introduced by LiveBot \cite{livebot} and other works \cite{zhang2023vcmaster, plvcg, 9533460, videoic}, focus on Chinese-language content sourced from platforms like Bilibili. Similarly, the MovieLC dataset \cite{knowledge} in Chinese-language is useful for structured movie-based interactions. To overcome these limitations—namely the predominance of Chinese-language content and limited video categories, we have constructed a large-scale, diverse, multimodal video-comments dataset in English-language.  Extracted from Twitch, this dataset covers 11 video categories, totaling $438$ hours and $3.2$ million comments. It is a valuable addition to the current live video comment datasets, as it is a real user-based dataset with pairs of videos and comments. Furthermore, with English-language content, it will have a broader global application and help researchers and practitioners with the advancement of live commenting technologies for English. It also introduces more challenges with diverse dialects and linguistic variations, including slang, gaming jargon, and informal, unstructured responses. Training models on such a complex dataset has been underexplored, establishing it as a demanding and impactful benchmark for understanding the semantic relationship between video streams and viewer conversations. Further, this dataset can be extended, in the future, to advance more challenging NLP research in informal language understanding and video-grounded multi-user conversational AI applications.

\section{Problem Statement}
Live Comment Generation in the context of videos involves generating relevant and coherent comments in real-time based on multiple input sources, including the video content and existing comments from viewers. 
It poses several challenges:

\begin{itemize}
  \item \textbf{Context Fusion}: Effectively combining information from the video content and existing comments to generate contextually appropriate comments.
  \item \textbf{Coherence with Existing Comments}: Ensuring that the generated comments are coherent with the sentiments and themes expressed in existing comments.
  \item \textbf{Diversity and Creativity}: Striving to generate diverse and creative comments.
\end{itemize}
Addressing these challenges will lead to the development of more sophisticated and effective Live Comment Generation systems.


Let \( \mathcal{V} \) represent the available video, and \( \mathcal{C} \) be the set of existing comments on that video. The goal is to generate a new comment \( \hat{r} \) that is contextually appropriate and informative.

Video \( \mathcal{V} \) can be represented as a sequence of frames \( \{f_1, f_2, ..., f_T\} \), where \( T \) is the total number of frames in the video. To capture the video context, we extract visual features \( \{v_1, v_2, ..., v_T\} \) using CLIP \cite{clip} or ResNet \cite{resnet} model, where \( v_i \) is a \( d_v \) dimensional vector representing the visual information from the \( i \)-th frame.


The existing comments \( C \) for a given video \( \mathcal{V} \) form a set of \( N \) comments, \( C = \{c_1, c_2, ..., c_N\} \), where \( c_i \) is the \( i \)-th comment. Each comment \( c_i \) is associated with a timestamp \( t_i \) that indicates the sequence in which the comment was posted during the video playback.

The goal of Live Comment Generation is to generate a new comment  \( \hat{r} \) that is contextually relevant to the visual content, as well as coherent with the existing comments. This can be formulated as finding the most probable comment given the video and the set of existing comments:

\begin{equation}
\hat{r} = \arg\max_{r} P( r | \mathcal{V}, C)
\end{equation}
where \( P(r | \mathcal{V}, C) \) represents conditional probability of generating comment \( r \) given the video content \( \mathcal{V} \) and existing comments \( C \).

An additional audio content modality is also captured in the dataset by converting the audio features into textual representations using a speech-to-text model \cite{whisper}. However, it was dropped from the main architecture as it did not contribute significantly to improving the model's performance.

\section{Semantic Frame Aggregation-based Transformer (SFAT) model}\label{transformer_model}
\begin{figure*}[htbp]
  \centering
  \includegraphics[scale=0.46]{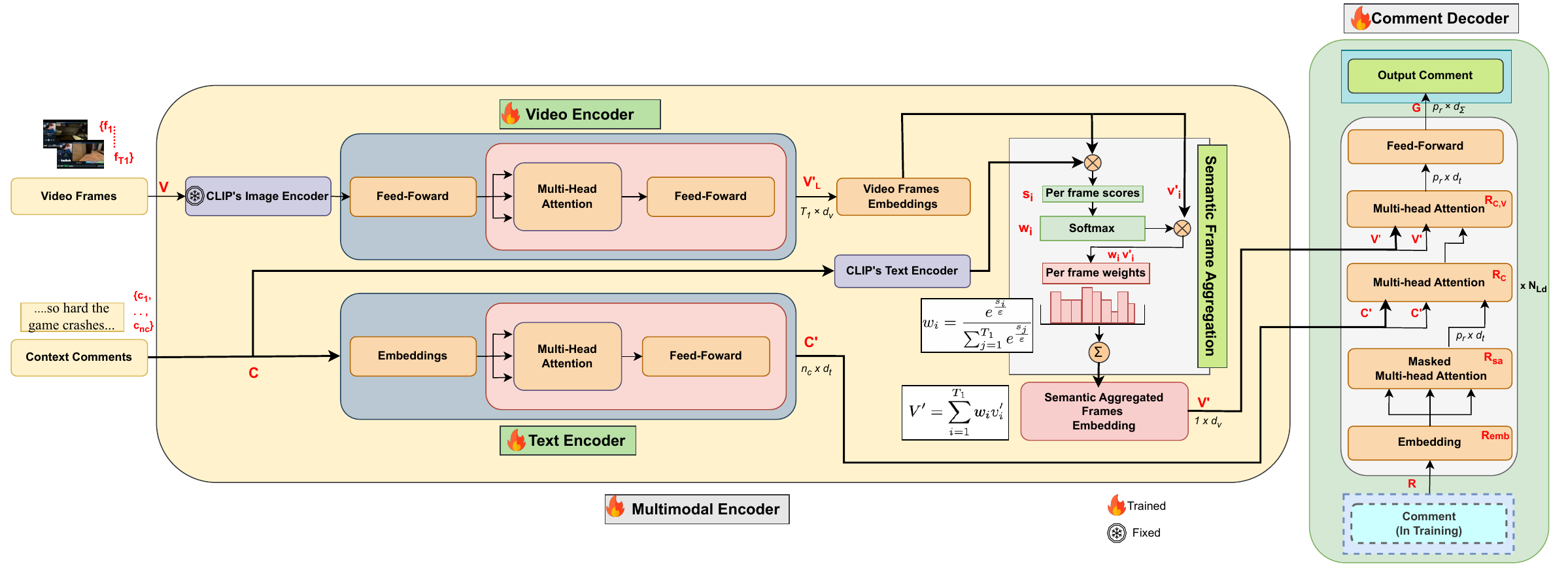}
  \caption{Semantic frame aggregation-based Transformer model for effectively extracting and combining visual and context comments modalities, and computing the weighted sum of frames for single video embedding. Here, \textcircled{x} represents dot-product operation and \(\Sigma\) represents aggregation.}
  \label{fig_mainarch}
\end{figure*}

We introduce our novel model, intricately designed to integrate video and context comments modalities for generating comments. Figure \ref{fig_mainarch} illustrates the details of the architecture for our proposed Semantic Frame Aggregation-based Transformer (SFAT) model. 

The proposed SFAT model not only extracts and combines visual and textual information effectively to generate comments, but also introduces a multimodal encoder. This encoder employs a sophisticated weighted sum of video frames to combine them into a single video embedding. It leverages CLIP's \cite{clip} visual-text multimodal knowledge by assigning weights to frames based on relevance to context comments for generating relevant and coherent comments.

\subsection{Multimodal Encoder}

Our model follows a transformer-based architecture in which we first encode each modality into a latent representation and then use a weighted sum of frames approach to predict a relevant output comment. Each modality is encoded through a Multi-Layer Transformer with $l_e$ layers and hidden dimension sizes $d_v$ and $d_t$ for video and text inputs respectively. Table \ref{tab:notation} shows a summary of the notations used.

\begin{table}[!t]
\caption{Mathematical Notations}
    \centering
    \scalebox{0.60}{
    \begin{tabular}{|c|c||c|c|}
    \hline
    Symbol & Definition & Symbol & Definition\\
    \hline
    $t$ & Starting timestep of the extract & $l_e$ & Layers in Transformer Encoder \\
    $T_1$ & Length of the context window & $l_d$ & Layers in Transformer Decoder\\
    $T_2$ & Length of the clip & $d_v$ & Hidden size of the Video Transformer Encoder\\
    $V$ & Video context vector & $d_t$ & Hidden size of the Text Transformer Encoder\\
    $C$ & Context comments vector &$p_c$ & Token count for context comments text features\\
    $V'_L$ & Latent video representation&$n_c$ & Number of context comments considered\\
    $R$ & Response vector&$p_r$ & Token count for response text features\\
    $V'$ & Aggregated frames representation &$d_\Sigma$ & Size of the Vocabulary\\
    $C'$ & Latent context comments representation & $G$ & Generated Response vector\\
    $s_i$ & Similarity score of video and context comments&$w_i$ &  Softmax of similarity score $s_i$\\
    \hline
\end{tabular}

}
    
    \label{tab:notation}
\end{table}

\vspace{0.4cm}

\textbf{Context Comments Encoder:} From the timestep $t$, we sample $n_c$ comments in the window $[t, t+T_1)$ where $T_1$ is the context window length, to produce the vector $C=\{c_1, \dots, c_{n_c}\}\in \mathbb{R}^{p_c\times n_c}$. Each comment $c_i$ goes through a word and positional embedding and is encoded by the BERT \cite{devlin-etal-2019-bert} based Comment Transformer Encoder to get the final representation of the comment. We use the CLS token $c'_i$ of the BERT-based embedding to get the representation of each comment $c_i$. Thus, we have a hidden representation of our context: 
$$\begin{aligned}
  C'&=\{c'_1, \dots, c'_{n_c}\}\\
    &= \{Transformer_c(c_i), i\in [1:n_c]\}
    \in \mathbb{R}^{n_c\times d_t} .
\end{aligned}$$

\textbf{Video Encoder:} Given a timestep $t$, we sample one frame per second in the window $[t, t+T_1)$ to obtain a vector $V=\{f_1, \dots, f_{T_1}\}$ of frames. Each frame is then encoded with a frozen CLIP \cite{clip} image encoder to obtain the representation $V_F=\{v_1, \dots, v_{T_1}\} \in \mathbb{R}^{T_1\times d_v}$. We then add positional embedding to each frame before passing them through the Transformer Encoder. The latent representation for video frames is:
$$\begin{aligned}
  V'_L&=\{v'_1, \dots, v'_{T_1}\}\\
    &= Transformer_v(V_F)
    \in \mathbb{R}^{T_1\times d_v} .
\end{aligned}$$


\subsection{Frame Aggregation-based Video Embedding}
It is observed in the collected dataset that the visual content may not always align well with ongoing user conversations, and contextual information from user dialogue has a greater influence on the generated comment, reflecting the viewer's interest. Therefore, it is required to selectively emphasize the key-frames which are semantically similar to ongoing conversations rather than treating all frames with equal importance as employed in previous works.
Our approach assigns weights to the frames embedding \(V'_L\) obtained from the video encoder based on similarity to context comments. It combines them efficiently into a single video embedding, focusing on the most relevant visual content when generating the next comment.

The video frames embedding \(V'_L\) obtained from the video encoder are combined to create a single video embedding by computing a similarity score \(s_i\) between visual and textual contents. The similarity score \(s_i\) is calculated as the dot product between the video frames embedding \(V'_L\) and context comments embeddings obtained using CLIP's \cite{clip} text encoder. The score \(s_i\) is then passed through a softmax with a temperature hyperparameter \(\varepsilon\) acting as the argmax operation to obtain the weighted sum of all the frames in the video.

The final aggregated frames embedding \(V'\) is represented below:
\begin{equation}
\begin{aligned}
    &V'= \sum_{i=1}^{T_1} w_i v'_i \in \mathbb{R}^{d_v},\\
    &\text{ where }  w_i = \frac{e^{\frac{s_i}{\varepsilon}}}{\sum_{j=1}^{T_1} e^{\frac{s_j}{\varepsilon}}} \in \mathbb{R}^{T_1\times n_c} .
\end{aligned}
\end{equation}
Here, \(v'_i\) is the normalized feature vector for each frame, and \(w_i\) represents the weight derived from softmax normalization of similarity scores \(s_i\) between video frames and context comments.

\subsection{Comment Decoder}
The decoder comprising of $l_d$ layers follows a modified transformer architecture with cross-attention mechanisms to attend to each modality, namely, video and context comments. Unlike standard Transformer decoders, our comment decoder incorporates modality-specific cross-attention layers that not only align the masked target comment with contextual comment embeddings from the text encoder but also integrate visual information from the aggregated video frame embeddings. This ensures that the generated comment reflects contextual cues from both chats and video. It takes the encoded representations of the context comments and the semantically aggregated video frames from the multimodal encoders. It uses them to generate the target comment selected from the response window $[t+T_1, t+T_2)$, denoted by $R=\{r_1, \dots, r_{p_r}\}$. 
\vspace{0.08cm}

\textbf{Target Comment Processing:} During training, the first input to decoder is the target comment $R$. It goes through a word and positional embedding layer, followed by a masked self multi-head attention layer to ensure that the decoder attends only to previously generated tokens.

$$\begin{aligned}
R_{emb} &= Embedding(R) ,\\
R_{sa} &= SelfAttention(R_{emb}) \in \mathbb{R}^{p_r \times d_t} .
\end{aligned}$$

The resulting processed target comment $R_{sa}$ serves as input for the subsequent cross-attention layers.

\textbf{Cross-Attention Layers:} The processed target comment is sequentially passed through two different cross-attention layers, each designed to attend to a specific modality.
\begin{align*}
    R_c &= CrossAttention_c(R_{sa}, C', C') , \\
    R_{c, v} &= CrossAttention_v(R_c, V',V')
       \in \mathbb{R}^{p_r \times d_t} .
\end{align*}

$CrossAttention_c$ layer aligns the masked target comment with the context comments embeddings from the text encoder, enabling the decoder to incorporate textual cues from the chat-context.

$CrossAttention_v$ layer incorporates information from the aggregated video frame embeddings, ensuring that the generated comment reflects the visual context.

\textbf{Prediction Layer:} After passing $R_{c, v}$ through a Feedforward layer, the prediction is done through a simple linear layer mapping the final hidden state to the vocabulary.
$$\begin{aligned}
G &= Linear(FeedForward(R_{c, v}))
  \in \mathbb{R}^{p_r \times d_\Sigma} .
\end{aligned}$$

This predicts the next token in the target comment, ensuring coherence with the multimodal input.

\subsection{Training Setup}
Our model's training process consists of two primary stages:
\begin{itemize}
\item \textbf{Pretraining:} In this stage, the text encoder for context comments is pretrained using a masked language modeling (MLM) task. Given a target text \( Y = \{y_1, \dots, y_n\} \), a masked version \(\widetilde{Y} = \{\widetilde{y_1}, \dots, \widetilde{y_n}\}\) is created where \(\forall i, \widetilde{y_i} = [MASK]\) with probability \( p \) and \(\widetilde{y_i} = y_i\) with probability \( 1-p \). The model is then trained to predict the original tokens from the masked ones using the cross-entropy loss. This process enables the text encoder to capture the underlying semantics and structure of the language used in context comments:
$$\begin{aligned}
\mathcal{L}_{\text{pretrain}} = -\mathbb{E}_{y_i \sim Y, y_i \neq \widetilde{y_i}} \left[\log \left(p({y_i}|\widetilde{y}_{\backslash i})\right) \right] .
\end{aligned}$$    
\item \textbf{Training:} In the second stage, the entire model, including the video and context comment encoders, as well as the decoder, is trained. The objective in this phase is to predict the target comment R from the multimodal inputs. The loss function is the cross-entropy loss applied to the entire sequence of target tokens:

$$\begin{aligned}
    \mathcal{L}_{\text{train}} = -\mathbb{E}_{r_i \sim R} \left[\log \left(p({r_i}|r_{< i})\right) \right] .
\end{aligned}$$

For every query, rather than having the model consistently predict a fixed target comment, we introduce randomness by selecting a comment uniformly at random from the response window. The model is then tasked with predicting this randomly chosen comment using the teacher-forcing technique. By not being bound to a single target for every query, the model is encouraged to generate a broader variety of comments, promoting richer and more diverse output during the generation phase.
\end{itemize}

\section{VideoChat Dataset}

\subsection{Data Collection and Preprocessing}
\begin{figure}[!t]
  \centering
  \includegraphics[width=0.40\textwidth]{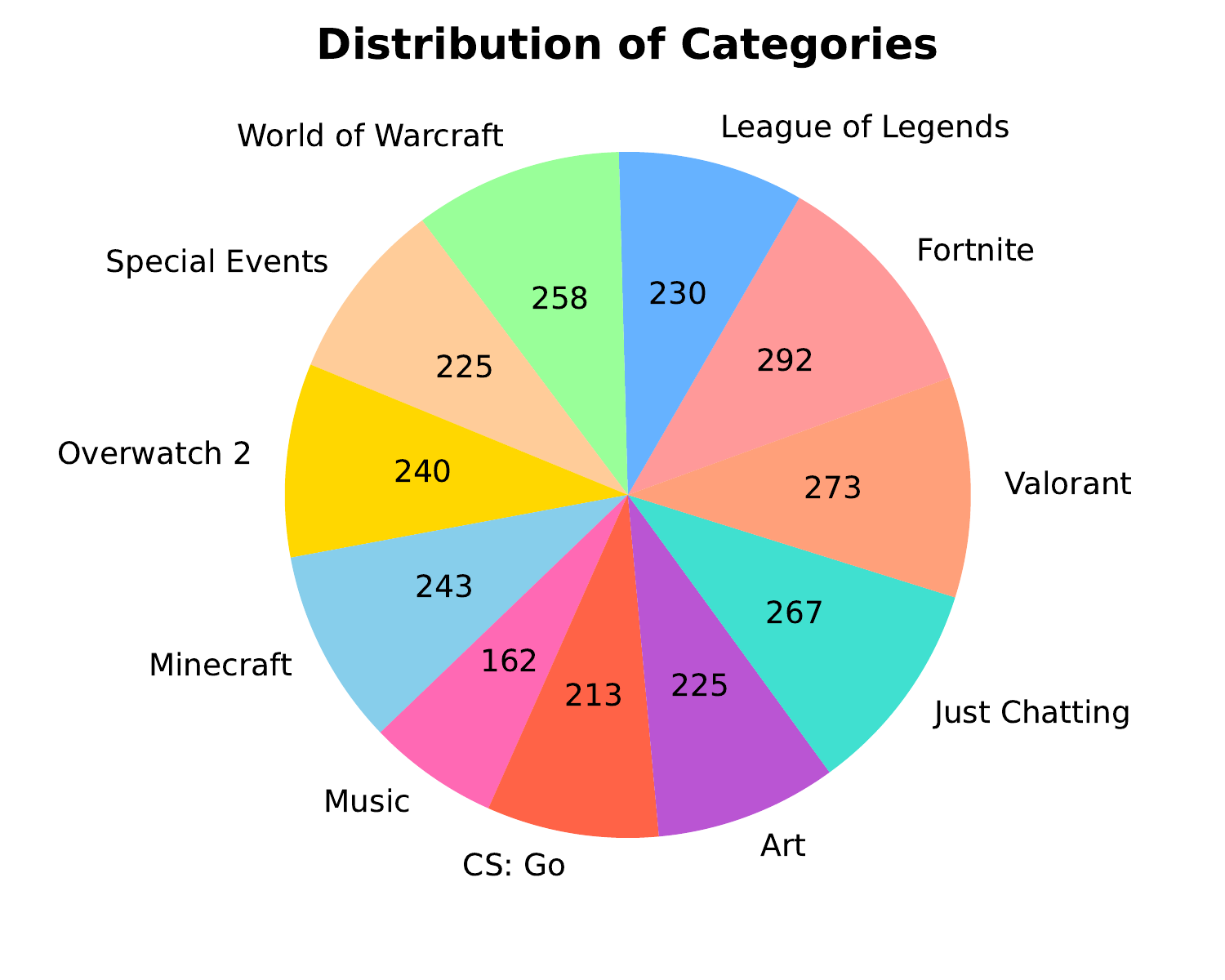}
  \caption{Repartition of extracted videos across the categories for VideoChat dataset.}
  \label{fig_category}
\end{figure}

The content that we propose in our dataset is extracted from the Twitch website. While Twitch's main audience is oriented towards video game content, we try to extend our dataset to a wide variety of content available on the platform. Figure \ref{fig_category} shows the number of videos per category extracted from Twitch to construct our VideoChat dataset. Half of the categories correspond to popular video games on the website, whereas the other half are related to a wide range of subjects such as music or arts which are categories that may invite the viewers to write longer and more coherent comments than the one we can usually see in video games category. We believe that expanding the range of categories will make the dataset applicable to a broad spectrum of tasks.


To collect the videos, we use the following process: using the TwitchAPI, we first find the top viewed categories and manually choose 11 of them to ensure the previously explained diversity. Then, we retrieve in each category the top viewed live streams of the week for four consecutive weeks and filter them to get at most two videos per week of the same streamer. Using the open-source tool TwitchDownloaderCLI, we download the entirety of the comments from these videos and finally use these comments to find the 30 minutes with the highest comment density in each video and download these parts with the same tool.

We then perform several transformations on these raw videos to enhance the dataset's usability, ensure anonymity, and reduce the storage requirements. Each video is first reduced from $1080p60$ to $720p5$ to conserve space while maintaining adequate quality and frame count for various tasks. Each frame is then encoded into a $512$-dimensional vector using the CLIP \cite{clip} image encoder, to anonymize the videos and facilitate computation. Some of our comparison methods also use the ResNet50 \cite{resnet} model pre-trained on ImageNet to obtain a vector of $2048$ dimensions for each video frame. Since most audio information in these videos corresponds to the streamer's speech, we employ a pretrained Speech2Text model \cite{whisper} to transcribe the content. Additionally, we remove usernames from the data to preserve privacy. 

Finally, each video is divided into $30$-second clips and sampled at $1$ frame per second for our task \cite{livebot, knowledge}. Each clip consists of a $20$-second context window and a $10$-second response window \cite{pasunuru-bansal-2018-game}.

\subsection{Dataset Analysis}
\label{dataset_analysis}

Our VideoChat dataset contains videos from $11$ categories and $575$ streamers for a total of $438$ hours and $3.2$ million comments. It is evaluated against four other major datasets in the domain, as detailed in the comparative statistics presented in Table II. Some additional information such as the dataset languages, number of video categories, collection websites, year of publication, and others, is also provided in Table \ref{tab:1}. 

Three of these datasets \cite{livebot, videoic, knowledge} constructed with Danmu comments are considered popular, but are in Chinese. The only other dataset available in English is Twitch-FIFA \cite{pasunuru-bansal-2018-game}. Compared to it \cite{pasunuru-bansal-2018-game}, our dataset offers a clear advantage in terms of dataset size, content variety, comment density, and overall usability. It encompasses a wide range of content and categories, capturing diverse English language user interactions and reactions. 
\begin{table}[!t]
 \caption{Comparison among different Live Video Commenting datasets}
    \centering
    \scalebox{0.70}{
    \begin{tabular}{|l|lllll|}
    \hline
    Dataset & LiveBot \cite{livebot} & VideoIC \cite{videoic} & MovieLC \cite{knowledge} & Twitch-FIFA \cite{pasunuru-bansal-2018-game} & VideoChat\\
    \hline
    \#Videos & 2,361 & 4,951 & 85 & 49 & 873\\
    \#Comments & 895,929 & 5,330,393 & 1,406,219 & 168,094 & 3,200,799\\
    Duration (h) & 114 & 557 & 175 & 86 & 438\\
    \hline
    \hline
    Avg. Duration (s) & 174 & 405 & 7,412 & 6,318 & 1,800\\
    Avg. \#Comments & 380 & 1,077 & 16,544 & 3,430 & 3,666\\
    Avg. \#Words & 5.42 & 5.39 & 6.53 & 5.55 & 4.02\\
    \hline
    Comment Density (c/s) & 2.18 & 2.66 & 2.23 & 0.54 & 2.02\\
    \hline
    \hline
    Comments Type & Danmu & Danmu & Danmu & Live chat & Live chat\\
    Language & Chinese & Chinese & Chinese & English & English\\
    \#Video Categories & 19 & 6 & 1 & 1 & 11\\
    Website & Bilibili & Bilibili & QQ & Twitch & Twitch\\
    Publication Year & 2019 & 2020 & 2023 & 2018 & ..\\
    \hline
\end{tabular}}
    \label{tab:1}
\end{table}

To the best of our knowledge, the VideoChat dataset stands as the largest and most diverse English video content dataset. The total duration of our dataset and our emphasis on segments with the highest comment density ensures that it is focused on the most engaging parts of the videos, enhancing its relevance for studying live English-language interactions. By incorporating these features, our dataset provides a substantial contribution to the field, particularly for researchers and practitioners focusing on the English-speaking live video interaction domain. To our knowledge, no other English live-video comment dataset is available after Twitch-FIFA \cite{pasunuru-bansal-2018-game}, published in the year 2018.


Moreover, we see that in live commenting, viewers frequently employ a diverse array of emotes, which cannot be handled as conventional text. To accommodate this, we have gathered and stored all the emotes used in the collected comments, resulting in a collection of $44,716$ different time-embedded emotes. 

The diversity in the language used by the viewers contributes to a rich vocabulary size within the dataset, encompassing $541,811$ unique words and an average of $4.02$ words per comment. However, owing to the real-time nature of the comments, it is not possible to guarantee grammatical accuracy or the use of extended sentences by users. This extensive vocabulary capturing various expressions and reactions might pose additional challenges for model training, but also offers a platform to evaluate the system's ability to understand and generate nuanced responses.

\section{Experiments}
\subsection{Evaluation metrics}\label{setup}
Comment generation task is a subjective and creativity-driven process that does not have a definitive correct answer. Video comments vary widely, making it intractable to find all references for comparison to model outputs \cite{livebot}. Consequently, existing generative evaluation metrics, such as BLEU \cite{papineni2002bleu} score or ROUGE \cite{lin2004rouge} score, are not suitable for this task. A retrieval-based evaluation method, sorting candidate comments based on the log-likelihood score, better aligns with the nature of this task \cite{livebot}.

Given a query (in our case, the video and comments contexts) and a set of candidates, retrieval metrics measure how well the model can retrieve or rank the relevant responses. Following the earlier works \cite{livebot, pasunuru-bansal-2018-game, knowledge, videoic}, we evaluate our model's performance using the below retrieval metrics:

\begin{itemize}
    \item \textbf{Recall@K:} Measures the percentage of times the true positive response is within the top \(K\) predictions. In our experiments, we compute Recall@$1$, Recall@$2$, and Recall@$5$.
    
    \item \textbf{Mean Rank (MR):} Average rank of the positive response among the list of candidate responses. Lower MR indicates better performance.
    
    \item \textbf{Mean Reciprocal Rank (MRR):} Average of the reciprocal ranks of the positive responses. For a particular query, if the rank of the true positive response is \( r \), the reciprocal rank is \( 1/r \). Higher MRR indicates better performance.
\end{itemize}

To construct different candidate sets, we select a list of 10 candidate responses from the entire dataset for each query. We adopt three  methodologies for the candidate selection:

\begin{enumerate}
    \item \textbf{Cosine Similarity:} Potential candidate comments are chosen based on their cosine similarity with the chat context. By using this metric, we ensure that the selected candidates are closely related and contextually relevant to the context comments.
    
    \item \textbf{Popularity:} Comments are selected based on their frequency in a live stream, reflecting the most prevalent reactions or sentiments within the community. This approach captures the commonly repeated comments that resonate with a larger audience.
    
    \item \textbf{Random:} Candidates are picked at random from the dataset, providing an understanding of the baseline performance of the retrieval mechanism.
\end{enumerate}
By employing these methods, we aim to comprehensively view the model's retrieval capabilities across varied selection criteria. Additionally, we also conducted a human evaluation to assess the relevance and appropriateness of the model's responses with human perception and judgment.

\subsection{Comparison Methods}
We evaluated our SFAT model's performance against the following methods: TTE model \cite{livebot} and its variants, Video-ChatGPT \cite{maaz2023video}, VideoIC \cite{videoic} and KLVCG \cite{knowledge}. The \textbf{Triple Transformer Encoder (TTE)} model is built upon the pioneering work of \textbf{LiveBot} \cite{livebot}, a representative work to address live video commenting. It utilizes a transformer-based architecture where each modality is encoded into a latent representation and passed to the decoder through different cross-attention layers to generate the output comment. The multimodal transformer employed by it to integrate video and textual contexts and its training objective aligns with our goal of generating contextually coherent comments by prioritizing the most relevant video frames. Moreover, its architecture inherently allows modifications to incorporate mechanisms such as frame weighting, making it an ideal foundation for building and evaluating our SFAT model. We further made some enhancements in the LiveBot \cite{livebot} model, such as using CLIP \cite{clip} embedding in place of ResNet \cite{resnet} for video-frames, dropping encoder-side attention, and adding audio modality.

\textbf{Video-ChatGPT} \cite{maaz2023video} is used as another method for comparison as we sought to evaluate the state-of-the-art multimodal LLM for the comment generation task. It integrates CLIP’s visual encoding with a language generation model (e.g., Vicuna \cite{chiang2023vicuna}), showcasing the power of LLMs in handling visual-text inputs. While Video-ChatGPT is designed for general video-understanding tasks (e.g., visual question answering and scene description), we extended its application to video commenting. To our knowledge, we are the first ones to employ a multimodal LLM for a video-comment generation task.  As a widely recognized method for multimodal tasks, it serves as a valuable benchmark to evaluate how SFAT performs in comparison. The experiments conducted on this multimodal LLM (discussed in detail in Sections \ref{main_result} and \ref{limits}) provide some valuable insights into its performance on this complex multimodal interactive task.

For our research objective of generating contextually coherent comments from given video-chat input, we also selected the VideoIC \cite{videoic} and KLVCG \cite{knowledge} models for evaluation and comparison on our dataset. Details of our comparison methods are discussed below:

\begin{enumerate}
    \item \textbf{TTE (ResNet):}  For the first comparison method, we use ResNet for encoding the visuals in the TTE model as used in LiveBot \cite{livebot}. It employs three separate encoders for each modality, namely, video-frames, context-comments and audio, followed by a decoder that generates the final comment based on multimodal inputs. The cross-attention mechanism between different modalities at the encoder side is dropped to see the effect on the results.

    \item \textbf{TTE (ResNet) with cross-attention:} The architecture is the same as the first comparison method using ResNet for encoding the visuals with added cross-attention mechanisms at the encoder side from context comments encoder to video and audio encoders, and from audio encoder to video encoder, respectively.
    
    \item \textbf{TTE (CLIP):} We use the same architecture as in the first TTE (Resnet) method but replace the Resnet visual embeddings with CLIP embeddings for improved visual-text alignment and semantic richness in feature representation.
    
    \item \textbf{TTE (CLIP) with cross-attention:} It uses CLIP embeddings for video frames with added cross-attention mechanisms at encoder side from context comments encoder to video and audio encoders, and from audio encoder to video encoder, similar to the second TTE comparison method.
    \item \textbf{Video-ChatGPT:} It is a video-conversation based multimodal LLM that merges the pretrained CLIP visual encoder with the Vicuna \cite{chiang2023vicuna} decoder to handle visual-text inputs.
    \item \textbf{Video-IC:} In its original form, VideoIC \cite{videoic} relies on both past (left) and future (right) contexts to predict temporal relations. Since in our generation task we only consider past chat-context for target time-step generation without accessing future context, we used the model's encoder-decoder based module for comment generation and adapted it to focus solely on learning temporal dependencies between past frames and target time-step.
    \item \textbf{KLVCG:} The original KLVCG \cite{knowledge} work proposed for movie dataset employs two types of external knowledge: the knowledge graph and the comments from other videos. To adapt a knowledge-enhanced model for our diverse dataset without any factual information, we used the model's TF-IDF based strategy for knowledge enhancement by the comments from other videos without relying on knowledge graphs.
    
\end{enumerate}
\begin{table*}[!t]
\caption{Evaluation results on the VideoChat Dataset for our proposed Semantic Frame Aggregation-based Transformer (SFAT) model, Triple Transformer Encoder (TTE) model \cite{livebot}, Video-ChatGPT \cite{maaz2023video}, VideoIC \cite{videoic} and KLVCG \cite{knowledge}. R@k represents Recall@k, MR is Mean Rank and MRR is Mean Reciprocal Rank. Highest scores are highlighted in bold, and second-highest scores are underlined.}
\resizebox{\textwidth}{!}{
 \begin{tabular}{|c|ccccc|ccccc|ccccc|}
\toprule
\multirow{2}{*}{\textbf{Model}}                                                              & \multicolumn{5}{c}{\textbf{Cosine Similarity}}                                                      & \multicolumn{5}{c}{\textbf{Popularity}}                                                                & \multicolumn{5}{c}{\textbf{Random}}                                                                 \\                                                                                                   & \textbf{R@1} & \textbf{R@2} & \textbf{R@5} & \textbf{MR} & \textbf{MRR} & \textbf{R@1} & \textbf{R@2} & \textbf{R@5} & \textbf{MR} & \textbf{MRR} & \textbf{R@1} & \textbf{R@2} & \textbf{R@5} & \textbf{MR} & \textbf{MRR} \\

\midrule
TTE (ResNet)                                                                 & 14.43               & \underline{27.46}               & 56.20               & \underline{5.06}               & \underline{0.34}         & 17.19               & 28.07               & 57.35               & 4.99               & 0.36         & 14.89               & 26.96               & 57.83               & 4.96               & \underline{0.35}        \\

\begin{tabular}[l]{@{}c@{}}TTE (ResNet) with cross-attention\end{tabular} & 14.13               & 26.77               & 55.22               & 5.11               & \underline{0.34}         & 18.67               & 30.44               & 60.72               & 4.77               & 0.38         & 15.44               & 27.40               & 56.53               & 4.99               & \underline{0.35}         \\

TTE (CLIP)                                                                   & 13.96               & 27.20               & 55.66               & \underline{5.06}               & \underline{0.34}         & 19.25               & 30.91               & 59.74               & 4.80               & 0.38         & 14.82               & 26.70               & 58.41               & 4.93               & \underline{0.35}  
\\

\begin{tabular}[l]{@{}l@{}}TTE (CLIP) with cross-attention\end{tabular}    & 13.83               & 25.81               & 56.09               & 5.09               & 0.33         & 16.48               & 27.13               & 56.37               & 5.03               & 0.35         & 14.89               & 27.16               & 57.87               & 4.95               & \underline{0.35}         \\
TTE (drop context comments)                                                                   & 11.53               & 22.73               & 53.27               & 5.28               & 0.31         & 15.82               & 25.87               & 54.29               & 5.18               & 0.34         & 11.62               & 22.82               & 54.18               & {5.24}               & {0.31}  
\\

TTE (drop audio modality)                                                                   & 14.48               & 26.68               & \underline{56.31}               & 5.07               & \underline{0.34}        & 19.77               & 31.02               & 61.65               & 4.65               & 0.39         & \underline{15.58}               & \underline{27.87}               & 57.91               & \underline{4.92}               & \underline{0.35}  
\\
\midrule
Video-ChatGPT                                                                                    & 11.05               & 23.32               & 55.69              & 5.15               & 0.32         & 21.39             & \underline{36.32}               & \underline{68.31}               & \underline{4.23}                & \underline{0.42}         & 11.94               & 24.07               & 58.97             & 4.97               & 0.33         \\

\midrule
Video-IC                                                                                  & 11.17               & 21.48               & 50.52              & 5.48               & 0.30         & 16.91             & 26.50               & 51.11               & 5.39                & 0.34         & 10.89               & 21.42               & 50.55             & 5.45               & 0.30         \\
\midrule
KLVCG                                                                                 & 12.99               & 24.04               & 55.41              & 5.15               & 0.33         & \textbf{28.33}               & \textbf{43.89}               & \textbf{71.25}               & \textbf{3.94}                & \textbf{0.47}         & 14.04               &  26.51               & \underline{59.36}               & 4.89               & 0.34         \\
\midrule
SFAT (all modalities)                                                                               & \underline{14.50}               & 26.68               & 56.18               & 5.07               & \underline{0.34}         & 17.12               & 28.52               & 57.76               & 5.00               & 0.36         & 14.78               & 26.22               & 57.19               & 5.01               & 0.34         \\
\textbf{SFAT (main model)} & \textbf{15.30} & \textbf{28.61} & \textbf{{58.11}} & \textbf{4.95} & \textbf{0.35} & \underline{22.25} & 35.56 & 64.86 & 4.45 & 0.41 & \textbf{16.67} & \textbf{28.76} & \textbf{59.58} & \textbf{4.81} & \textbf{0.36} \\

\midrule
\end{tabular}}
    \label{tab:results1}
\end{table*}

\subsection{Ablations Study}
We conducted some ablation studies to assess how different modalities impact model performance. We dropped the context comments modality to see the impact on results. Similarly, we also dropped the audio modality to study its contribution. The results are reported in Table \ref{tab:results1} as TTE (drop context comments) and TTE (drop audio modality). As evident from the results, the audio modality does not significantly enhance the model's performance (Recall@1 remains $14.48$ for the Cosine-similarity candidate set). Whereas, the context-comments modality plays a crucial role (Recall@1 drops to $11.53$ when this modality is removed for the Cosine-similarity candidate set). Further, we integrated the weighted sum of the frames module to the TTE model with audio modality to see the impact on results. The findings are reported in Table \ref{tab:results1} as SFAT (all modalities). The results show that the performance improves markedly across all candidate sets in SFAT (main model) with audio modality dropped, compared to SFAT (all modalities). Details of the impact of modalities are further discussed in Section \ref{main_result}.

\subsection{Experimental settings}
For the optimal configuration of our SFAT model and transformer-based TTE method with its variants, we set the learning rate to $10^{-4}$, batch size to $32$, pretrained the model for $100$ epochs and trained it for $200$ epochs. The Transformer encoder and decoder were configured with $4$ layers, $8$ attention heads and $256$ hidden dimensions. We set the dropout to $0.1$ and choose $5$ and $15$ context comments during training and evaluation respectively. 

The Video-ChatGPT \cite{maaz2023video} model was trained for $2$ epochs with a batch size of $8$. It employed the Adam optimizer with a learning rate of $2e-4$, a warmup ratio of $0.03$, and the temperature parameter set to 0.8. A cosine scheduler was applied as the learning rate scheduler for training, and the base model utilized is LLaVA-7B-Lightning-v1. For the VideoIC model, we employed the Adam optimizer with a learning rate of 0.0001 and a weight decay of $0.001$, trained for $100$ epochs (optimized at $25$) with a batch size of $32$. The model architecture featured $256$-dimensional embeddings and hidden states, $8$ attention heads, and 6 transformer blocks, with a maximum sequence length of $10$ tokens and a dropout rate of $0.1$. The temporal context was modified to include only preceding comments and zero following comments as our dataset featured only preceding comments. For the KLVCG model, we trained it with a learning rate of $10^{-4}$, learning rate decay of $0.9$, learning rate coefficient as $0.1$, batch-size as $128$ and external knowledge context-length of $120$, obtained by TF-IDF based comments retrieval from other videos, for $100$ epochs.

\textbf{Note on inference-time overhead:} The model once trained, enables near-instant inference for a test video sample. The trained model inially takes less than $4$ seconds to load on NVIDIA A100 80G server. With the available pretrained CLIP embeddings for video-clips, the trained model takes about $344.44$ seconds for generation of $2000$ samples with an average time of $0.17$ seconds per sample and total GPU memory usage of $2790$ MiB.




\subsection{Performance Comparison with Existing Methods}\label{main_result}
In Table \ref{tab:results1}, we present the performance of our SFAT model against other comparison models: TTE \cite{livebot} model and its variants, Video-ChatGPT \cite{maaz2023video}, VideoIC \cite{videoic} and KLVCG \cite{knowledge} across different metrics under various candidate selection methods. It provides insights into how different selection methods impact the model's retrieval capabilities. We also provide additional experimental results to study contribution of different modalities, on the overall performance of the models.

From Table \ref{tab:results1}, we can see that our main SFAT model performs well across all metrics, notably for the Popularity candidate set where Recall@1 increased from $19.77$ to $22.25$, Recall@2 from $31.02$ to $35.56$, Recall@5 from $61.65$ to $64.86$ and MRR from $0.39$ to $0.41$, compared to the Triple Transformer Encoder (TTE) models. The Random candidate set also showed a significant increase in Recall@1 from $15.58$ to $16.67$ and Recall@5 from $57.91$ to $59.58$, compared to the TTE models. 
%



    



Moreover, SFAT performs better than all other models for the Cosine-similarity and Random candidate sets. For the Cosine-similarity candidate set, comments closely related and contextually relevant to the chat sequence are selected. Due to their similarity to the chat context, the model assigns them a high log-likelihood score, often very close to the ground truth, causing the ground truth to not always appear in the top-k ranks when sorted by log-likelihood scores. Conversely, comments in the Randomness and Popularity sets, selected randomly or by frequency, do not match the chat context very closely and hence would get lower scores, causing the ground truth to appear in top-k ranks more frequently. Thus, the model performs better on the Popularity and Randomness sets than on the Cosine-similarity set.

Video-ChatGPT's \cite{maaz2023video} performance is notably higher for Popular candidate comments as it is adept at capturing frequently occurring patterns in popular comments. However, the model's generalization capabilities do not align well for the Cosine-similarity candidate set where very specific or less common contextual cues can be present in text. Similarly, it struggles to apply learned patterns effectively for the Random candidate set. Both these methods challenge the model's capability to apply generalized knowledge effectively for the high specificity required by Cosine-similarity and the unpredictability introduced by the Random candidate selection method. Here, our SFAT model with the weighted sum of frames approach provides better contextualization, outperforming Video-ChatGPT for semantically rich and specific outputs.

KLVCG \cite{knowledge} recalls knowledge and comments from other videos to enrich its inputs. The knowledge-enhanced model combines both local context from the immediate chat-context and video frames and external knowledge from similar comments in other videos. With tailored domain-specific knowledge enrichment mechanism, it performs better than Video-ChatGPT for the Popularity candidate-set. However, this external knowledge may dilute the model's ability to focus on the immediate context (e.g., unique visual or textual cues) of the target video. Hence, it is not performing so well for the Cosine-similarity candidate-set due to potential noise from less relevant content added during knowledge enrichment. In the future, we can integrate it with a better knowledge-filtering mechanism or dynamic weighting between the immediate context (current video and chat) and external knowledge for improved performance.

The VideoIC \cite{videoic} model shows relatively low performance compared to other methods as it is designed for tasks that rely on temporal relationships between past and future contexts. Without future context, the temporal prediction mechanism loses its effectiveness, leading to degraded performance for our dataset.

We also observe from the results that audio modality does not make a substantial contribution to improving the model's performance. This may arise from the audio content not aligning with the visuals and contextual dialogue in the dataset. It hinders the model's ability to improve generation quality by effectively integrating visual, audio, and text correlations. Dropping audio modality from the SFAT (all modalities) improves the results significantly. In contrast, context comments modality plays a significant role, as removing it from the model significantly drops scores across all metrics. Hence, in our main SFAT model, we drop audio modality and only utilize context comments modality with video content.

  
  
  

\begin{figure*}[!t]
  \subfloat[]{\includegraphics[width=0.52\textwidth]{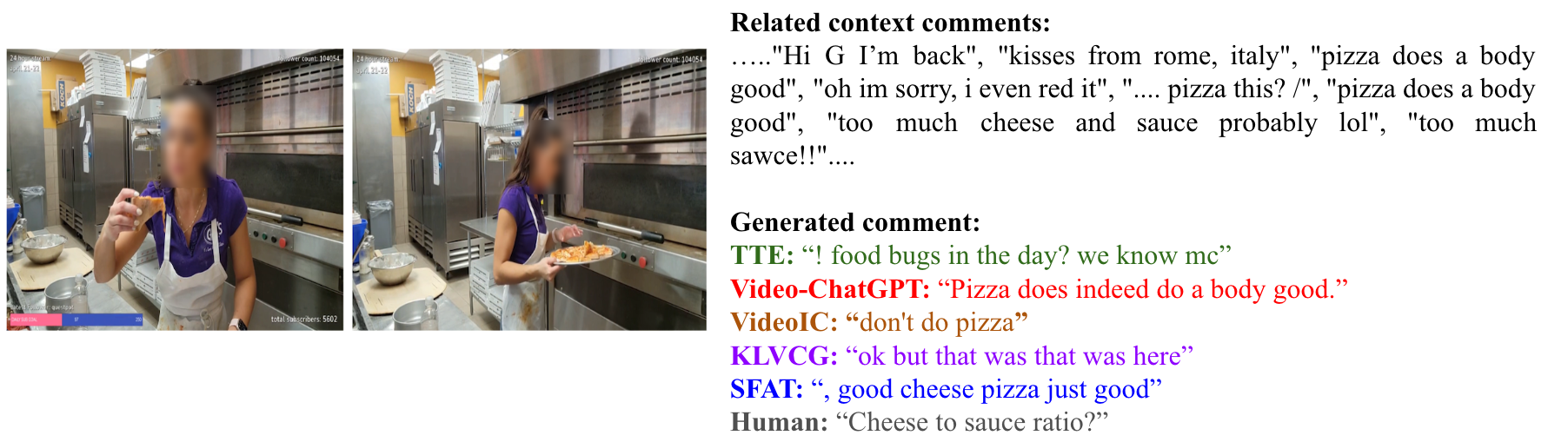}}
  \hspace{0.005\textwidth} 
  \subfloat[]{\includegraphics[width=0.52\textwidth]{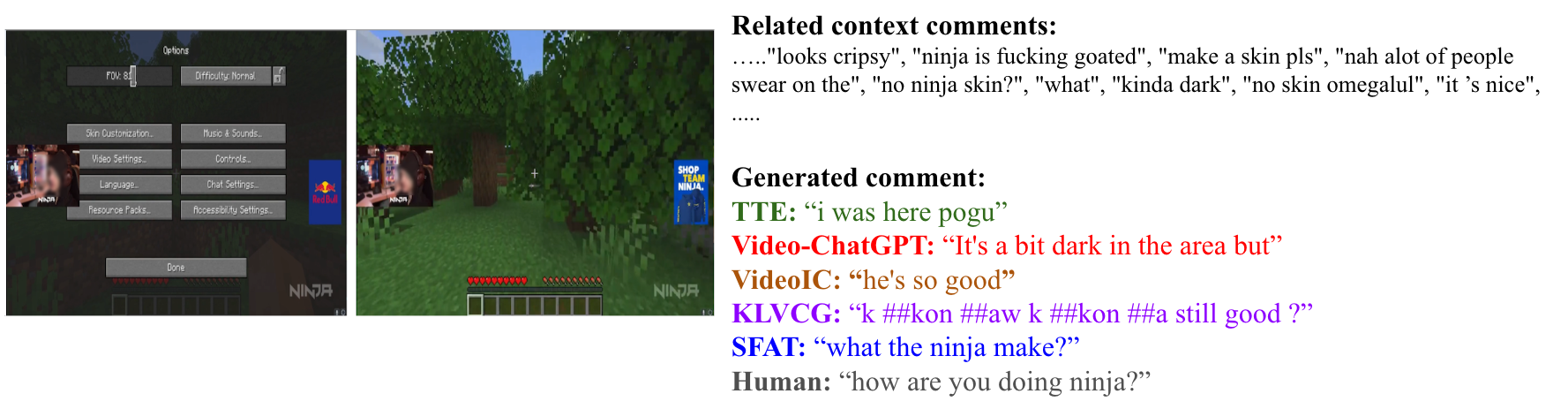}} \\
  
  \subfloat[]{\includegraphics[width=0.52\textwidth]{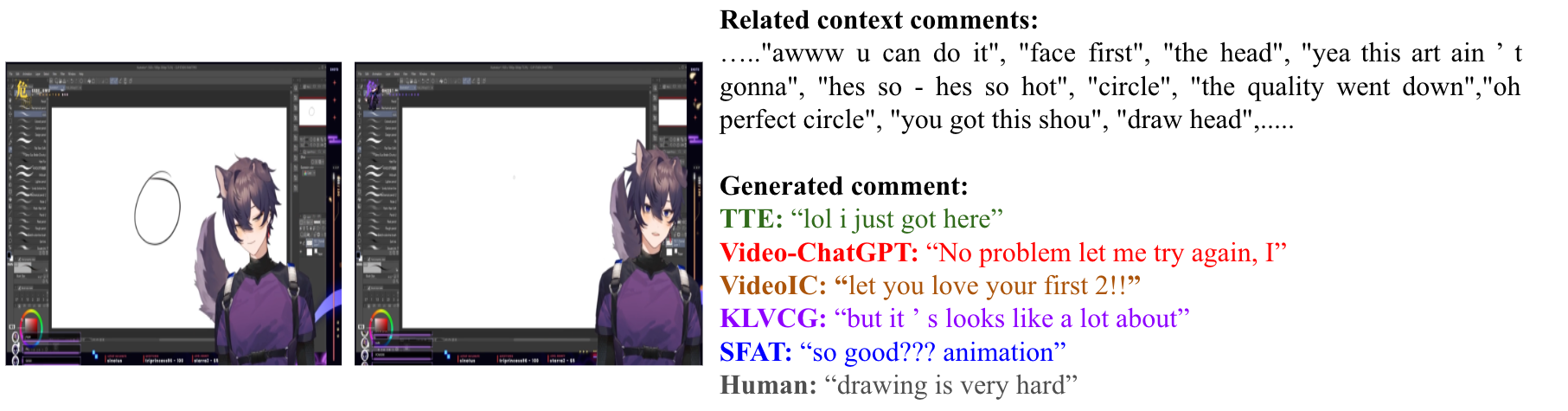}}
  \hspace{0.005\textwidth}
  \subfloat[]{\includegraphics[width=0.52\textwidth]{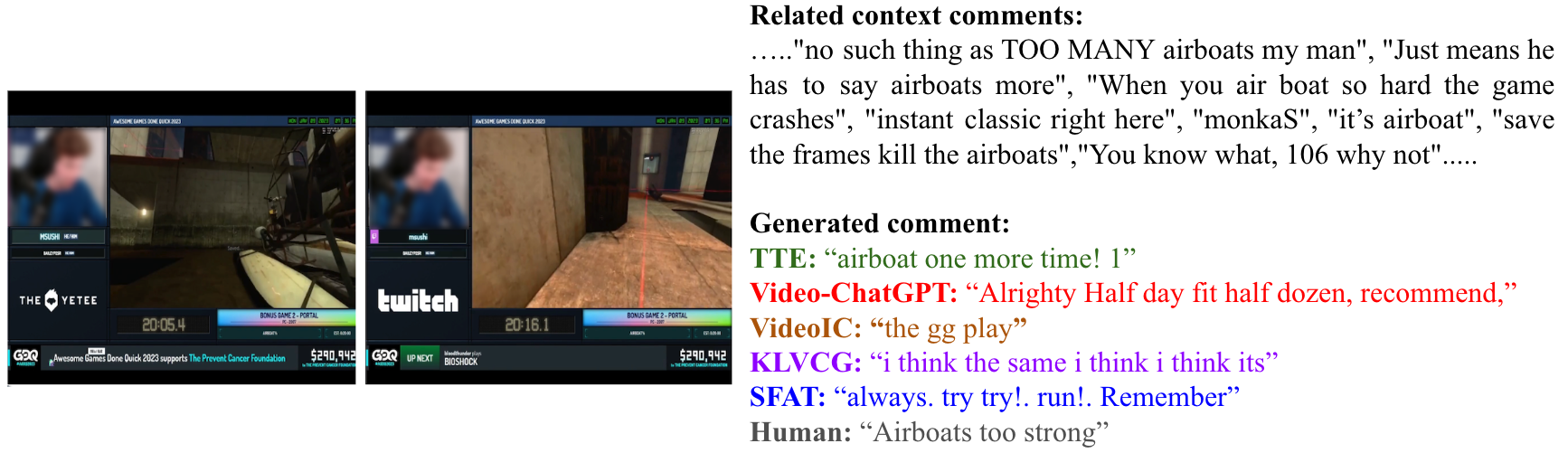}} \\
  
  \subfloat[]{\includegraphics[width=0.52\textwidth]{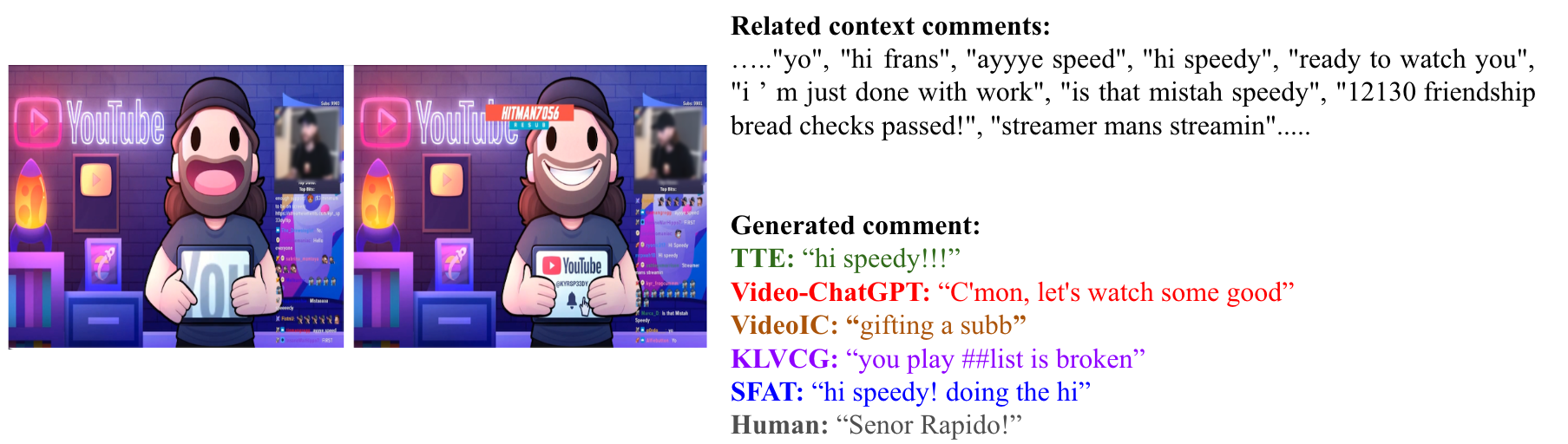}}
  \hspace{0.005\textwidth}
  \subfloat[]{\includegraphics[width=0.52\textwidth]{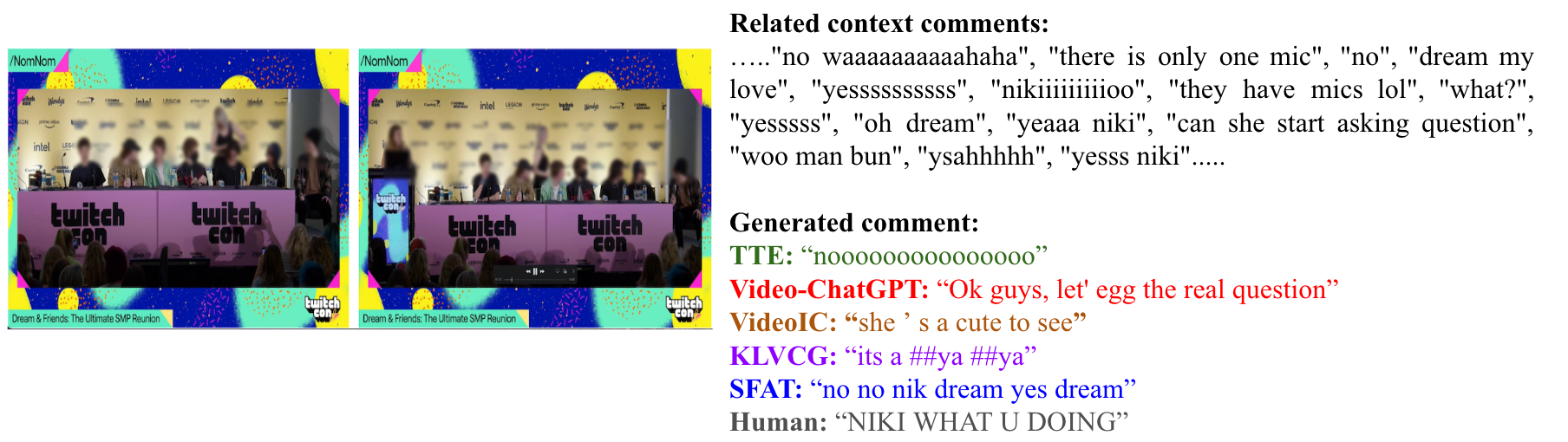}}
  
  \caption{
    Examples of video paired with related context comments (in black) and the generated comment for human evaluation on our proposed SFAT model (in blue) and the comparison methods: TTE (in green), Video-ChatGPT (in red), VideoIC (in brown) and KLVCG (in purple). Human-posted comments are shown in grey.
  }
  \label{fig:examplevideo}
\end{figure*}

\subsection{Human Evaluation}

Generating live comments from a given video is a subjective, creativity-driven process with no definitive correct answer. When evaluating creative comments, it is crucial to recognize that the goal is not to replicate a single ground truth reference. In some cases, a reasonable comment may be unfairly penalized simply because it deviates from the ground truth, even though it effectively aligns with the context of the live video. Consequently, such metrics may not fully capture the true performance or effectiveness for evaluating any method. To address this limitation, we additionally conducted a human evaluation to assess the relevance, fluency, and correctness of the generated comments.

Figure \ref{fig:examplevideo} shows some samples of the generated comments selected from different video categories for human evaluation. Given the input video sample and related context comments, we asked five human annotators to score the responses of the five models (TTE, Video-ChatGPT, VideoIC, KLVCG and SFAT ) and the human-posted comments based on three aspects: \textbf{Fluency}, \textbf{Relevance}, and \textbf{Correctness} \cite{livebot}. Fluency is intended to evaluate whether the generated comments are fluent regardless of their relevance to the videos. Relevance focuses on assessing the relevance between the generated comment with the context comments and the video content. Correctness aims to assess how confident we can be that the generated live comments look close to human-generated in the context of the video. For each of the three aspects, we require an integer between 1 and 5 as score, with higher scores indicating better performance. The final results are then calculated by averaging the ratings from the $15$ human annotators.

The examples shown in Figure \ref{fig:examplevideo} highlight the challenging task of correlating video frames with viewer comments. SFAT attempts to generate comments that align with the context comments and video content. For instance, in the first example, it generates comments about the ``cheese pizza" discussed in chat-contexts and visible in the video-frame. The nuanced understanding of both the visual content and the context aids in generating dynamic responses, capturing the essence of the conversation. In other examples (b) and (f), it generates a comment about the ``ninja" avatar and ``niki" being discussed in the chat-context. As discussed earlier, certain comments (for samples (b) and (f)) generated by KLVCG can contain some irrelevant content from external knowledge. Overall, we see that, compared to other models, SFAT generates context-aware comment that resonate with the viewer's interest, instead of a generic response.

\begin{table}[!t]
\caption{Human evaluation metrics results on the test-set samples (higher is better).}
\centering
 \begin{tabular}{|l||c|c|c|}
\hline
\textbf{Model} & \textbf{Fluency} & \textbf{Relevance} & \textbf{Correctness} \\
\hline
TTE & 3.11 & 2.73 & 2.87 \\
Video-ChatGPT & \textbf{3.35} & 2.85 & 2.75 \\
VideoIC & 3.15 & 2.96 & 3.02 \\
KLVCG & 2.14 & 2.16 & 2.12 \\
SFAT & 3.33 & \textbf{3.92} & \textbf{3.83} \\
\hline
Human & 3.81 & 3.97 & 4.03 \\
\hline
\end{tabular}

    \label{tab:humaneval}
\end{table}

Table \ref{tab:humaneval} presents the results of the human evaluation. Our model performs better than the other comparison methods: TTE, Video-ChatGPT, VideoIC and KLVCG, for Relevancy and Correctness metrics as it prioritizes relevant frames than treating all frames with equal importance. Video-ChatGPT performs slightly better than our model in Fluency metrics because, being an LLM-based model, it has been trained on a large human-language corpus, which enhances its linguistic fluency. We also assess the reference comments in the test set, which are posted by humans. The results indicate that these human-posted comments generally score good across all metrics, particularly for Correctness. Live-video comments are typically made in real-time environment and reflect immediate, unfiltered reactions. Viewers might respond to specific moments in the video, interact with other users, or even discuss unrelated topics, leading to a wide range of content in the comment section, often in informal language. Due to the spontaneity and diversity of live interactions, they may not always be very fluent or directly related to the video content, making relevant comment generation a challenging task. Compared to other methods, the Relevance score of our model is very close to the human-posted comment, demonstrating its ability to generate improved contextually relevant comment using semantic frame-aggregation mechanism.

\subsection{Evaluation on LiveBot Chinese Dataset}
The primary goal of this work is to prioritize relevant frames from ongoing conversations to generate contextually aligned comments and build a diverse multimodal video-commenting dataset in a global language such as English. While we do not aim to develop a language-agnostic model, the proposed semantic weighted-sum approach can be extended to datasets in other languages with appropriate modifications. To train our model on the existing Chinese dataset, the following changes are required: using Chinese BERT embeddings for text and a Chinese-Clip model for visuals, followed by the semantically aggregated frame-embeddings mechanism. We evaluated our SFAT model on the LiveBot dataset using available processed data with ResNet embeddings and without MLM-based pretraining. We obtained following results on LiveBot Chinese dataset: \( Recall@1 \): $17.41$, \( Recall@2 \): $31.74$, \( Recall@5 \): $67.99$, \( MR \): $4.44$ and \( MRR \): $0.39$ using candidate-set of 10. The results are comparable with our model's performance (in Table \ref{tab:results1}) on the English VideoChat dataset, demonstrating its versatility and adaptability for content beyond the English language.

\section{Limitations and Future Work}\label{limits}
Overall, live video comment generation is a relatively complex task requiring integrating video modality with multi-user comments. Our current work has some limitations, and we would like to address future research directions. 

Multi-user comments can be ambiguous and semantically diverse, as users may react differently to video events, respond to previous comments, or engage in dialogues, sometimes independent of the live-streaming. Due to the intricate multimodal context and ambiguous content, there are some inherent limitations to improving the model's performance beyond a certain score. 
The task becomes more challenging for English due to diverse global accents and dialects. Using slang, abbreviations, and gaming lingo in viewer conversations presents challenges for the model in accurately interpreting their meanings. It can also be difficult for the audio-to-text model to make accurate transcriptions. Another challenge in the dataset is that conversations sometimes diverge from the video content and focus more on ongoing dialogue. This makes it difficult for the model to correlate visual and textual elements effectively, leading to suboptimal performance without the weighted frames approach.

Moving forward, we aim to refine our dataset and enhance our model to handle better the nuanced demands of live comment generation across code-mixed languages and other multi-lingual datasets, enriching viewer engagement. Future enhancements in the transformer-based model will focus on optimizing attention mechanisms for the three modalities and improved integration techniques, capturing the nuances of emotes, slang, and informal expressions. Further, we will enhance our VideoChat dataset quality by adding more contextual details about the video, increasing the average comment length, and incorporating temporal sequences to capture conversations over time. 

Most of the works in multimodal LLMs \cite{maaz2023video,li2023llama} have focused on video-centric tasks such as visual question answering or video understanding with predefined queries and structured inputs. However, there is scope for exploring their potential in other complex multimodal interactive tasks, such as live video commenting, involving diverse and informal user-generated content.

\section{Conclusion}
In this research, we embarked on the intricate task of video comment generation in relevance to ongoing viewer conversations. By introducing the Semantic Frame Aggregation-based Transformer (SFAT) model, our work effectively addresses the overlooked problem of prioritizing relevant video frames in alignment with ongoing viewer interactions. This prioritization ensures the generation of comments that are more coherent and engaging compared to traditional approaches that treat all video frames equally.

Additionally, while previous datasets were limited to Chinese-language content, we developed a large-scale, diverse, multimodal English video comments dataset. It will cater to a wider global audience, supporting researchers and practitioners in advancing video-commenting technologies for English content. Furthermore, we conducted extensive quantitative experiments and human evaluation on the existing methods, along with a state-of-the-art multimodal LLM, to demonstrate the effectiveness of the SFAT model in utilizing frame prioritization to generate contextually relevant comments. We also demonstrated the adaptability of our semantic frame aggregation approach to Chinese-language content, highlighting its versatility beyond the English dataset.
The insights gained have deepened our understanding of the complexities and limitations involved in this challenging task, thereby paving the way for future research in developing more advanced live commenting systems.

\section{Ethical Considerations}
To mitigate the ethical risks associated with generating fake video comments, when employed on real-time platforms, a label can be included to distinguish AI-generated comments from human-posted ones, ensuring that viewers are aware of the source of the comments. Moreover, our current work focuses on benign video categories such as video games, music, and arts, primarily focusing on user engagement, where ethical risks are generally less severe. However, stricter oversight will be required when applying this methodology to more serious content that influences public opinion, education, or other critical areas. Therefore, to ensure its ethical use, we will restrict the usage of our method and make our model available to the AI community, contributing to the responsible development of AI-generated comments from video content.  

\section{Acknowledgement}
Anam Fatima has been supported by the Center for Design and New Media (A TCS Foundation Initiative supported by Tata Consultancy Services) at IIIT-Delhi and partially funded for this work by the NII International Internship Program 2023, Tokyo, Japan. We acknowledge Arnesh Batra's valuable support in benchmarking the performance of our model.

\bibliographystyle{IEEEtran}
\bibliography{mybibfile}  

\newpage

\vspace{11pt}

\vfill

\end{document}